\documentclass[11pt]{article}
\usepackage[preprint]{acl}
\usepackage{times}
\usepackage{latexsym}
\usepackage[T1]{fontenc}
\usepackage[utf8]{inputenc}
\usepackage{microtype}
\usepackage{inconsolata}
\usepackage{marvosym}
\usepackage{fontawesome5}
\usepackage{graphicx}
\usepackage{flafter}
\usepackage{xcolor}
\usepackage{tcolorbox}
\usepackage{listings}
\usepackage{amsmath}
\usepackage{booktabs}
\usepackage{multirow}
\usepackage{makecell}
\usepackage[table]{xcolor}
\tcbuselibrary{skins}
\definecolor{SpecViolet}{HTML}{64469A}
\definecolor{SpecLight}{HTML}{F0EAF8}
\definecolor{SpecCompactFill}{HTML}{F0EAF8}

\newtcolorbox{observationbox}[1]{
    colback=SpecLight,
    colframe=black,
    coltext=black,
    boxrule=0.5pt,
    arc=1mm,
    boxsep=0pt,
    left=6pt,
    right=6pt,
    top=5pt,
    bottom=5pt,
    before skip=6pt,
    after skip=6pt,
    before upper={\textbf{Observation.}\enspace}
}

\lstdefinestyle{messages}{
  basicstyle=\ttfamily\footnotesize,
  backgroundcolor=\color{black!3},
  rulecolor=\color{black},
  frame=single,
  framerule=0.5pt,
  framesep=6pt,
  columns=fullflexible,
  keepspaces=true,
  breaklines=true,
  showstringspaces=false,
  aboveskip=0pt,
  belowskip=0pt
}

\definecolor{CIGray}{HTML}{777777}
\newcommand{\CI}[2]{%
  \hspace{0.10em}\textsubscript{%
    \normalfont\fontsize{6.5}{7}\selectfont
    \textcolor{CIGray}{[#1,#2]}%
  }%
}

\newcommand{\promptvar}[1]{%
  \textcolor{SpecViolet}{\texttt{\{\detokenize{#1}\}}}%
}

\newcommand{\slotname}[1]{%
  \textbf{\texttt{\detokenize{#1}}:}%
}

\newtcolorbox{promptbox}[1]{
  enhanced,
  title={#1},
  colback=SpecLight!45!white,
  colframe=black,
  colbacktitle=SpecViolet,
  coltitle=white,
  coltext=black,
  fonttitle=\bfseries\normalsize,
  fontupper=\small,
  boxrule=0.65pt,
  titlerule=0pt,
  arc=1.8mm,
  outer arc=1.8mm,
  boxsep=0pt,
  left=8pt,
  right=8pt,
  top=6pt,
  bottom=6pt,
  toptitle=4pt,
  bottomtitle=4pt,
  before skip=0pt,
  after skip=0pt,
  before upper={\parindent=0pt\parskip=4.5pt\relax}
}

\title{SpecCompact: Accelerating LLM Agent Context Compaction with Slot-Parallel Semantic Speculative Verification}

\author{Yulu An\thanks{Equal contribution.} \\
  New York University \\
  \texttt{ya2191@nyu.edu} \\\And
  Yifei Feng\footnotemark[1] \\
  New York University \\
  \texttt{yf3005@nyu.edu} \\\And 
  Xiangyang Yin\footnotemark[1]\thanks{Corresponding author.} \\
  New York University \\
  \texttt{shawn.yin@nyu.edu} \\
  \AND
  \ \\[-1.88em]
{\normalfont\href{https://github.com/shawnyin128/Speculative-Compaction}{\faGithub\enspace SpecCompact}}
  }

\begin{document}
\maketitle

\begin{abstract}
LLM agents use context compaction to manage growing interaction histories, but summarizing long histories with large models can introduce substantial delays. Compaction therefore requires balancing information retention against latency. We observe that structured summary slots can retain information when generated independently. Based on this observation, we present SpecCompact, a framework that accelerates compaction through slot-parallel semantic speculation. SpecCompact fine-tunes a small drafter and generates slots in parallel. The target model verifies each proposed line and repairs rejected content, while subsequent drafting uses the updated slot prefix. System optimizations further reduce repeated context processing and model calls. Across AppWorld and $\tau^3$ benchmarks with two target models, SpecCompact achieves $2.04$--$5.63\times$ compaction speedups over Hermes-style summarization. Separate downstream evaluations reveal model- and task-dependent performance trade-offs. A user study within Hermes further shows an $8.68\times$ median compaction speedup.
\end{abstract}

\begin{figure}[!t]
    \centering
    \includegraphics[width=\columnwidth]{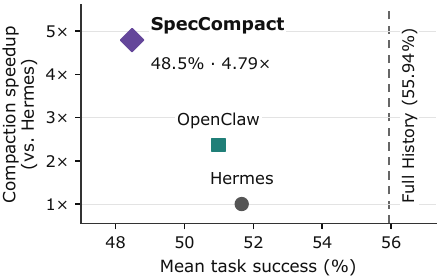}
    \caption{
        Task success and compaction speedup with Gemma4-31B-IT,
        measured in separate evaluations.
    }
    \label{fig:performance_speedup}
    \par\vspace{-15pt}
\end{figure}

\begin{figure*}[t]
    \centering
    \begin{minipage}[t]{0.48\textwidth}
        \centering
        \includegraphics[width=\linewidth]{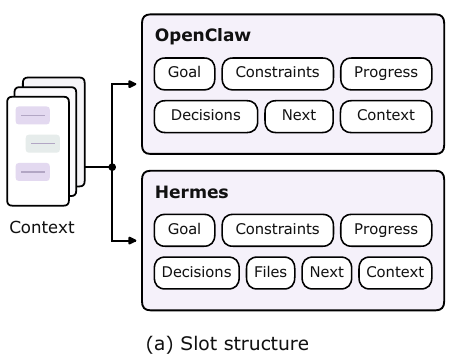}
    \end{minipage}\hfill
    \begin{minipage}[t]{0.48\textwidth}
        \centering
        \includegraphics[width=\linewidth]{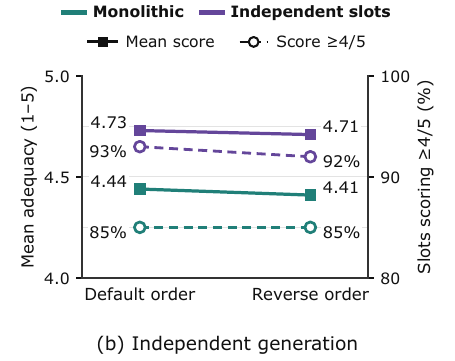}
    \end{minipage}
    \caption{Slot structure and independent generation.
    (a) Fields in OpenClaw and Hermes compaction templates.
    (b) Reference adequacy over 100 trajectories on LiveClawBench~\citep{long2026liveclawbenchbenchmarkingllmagents}, with six slots per trajectory.}
    \label{fig:slot-motivation}
    \par\vspace{-10pt}
\end{figure*}

\section{Introduction}
Large language model (LLM) agents now handle a growing range of tasks, especially personal work and coding~\citep{muse,dotagent,claudecode,codex}. While longer context windows have enabled better long-horizon performance, they also allow agents to accumulate larger histories of user requests, tool outputs, and intermediate decisions. Agent systems typically use context compaction to shorten these histories while retaining information needed for later actions~\citep{openclaw,hermesagent}. However, compaction itself requires processing a long context and decoding a summary token by token. The resulting delay can significantly degrade the user experience. Compaction therefore requires balancing information retention against the latency of producing the summary.

Token-level speculative decoding, a general method for accelerating decoding, can speed up this compaction process. Specifically, a smaller model with lower inference costs can accelerate generation while preserving the target model's output distribution by target model verification~\citep{leviathan2023,chen2023specsampling}. However, valid summaries can differ in wording yet have the same effect on later actions. Thus, although speculative decoding preserves lossless compaction behavior, it may not fully exploit this flexibility. Recent semantic speculation methods use target-model judgments to accept sentence-level drafts~\citep{specreason,dreamr}, making them better suited to compaction. However, applying these methods to a single summary still requires sequential generation, which can be slow for long contexts. Moreover, asynchronous compaction overlaps summarization with continued agent execution, while parallel compaction summarizes input blocks concurrently~\citep{slipstream,parallelcompaction}. These approaches raise a further question: \textit{can the structure of the summary itself guide parallel generation and target-model verification?}

Existing compaction templates provide a useful starting point. Agent frameworks such as OpenClaw and Hermes~\citep{openclaw,hermesagent} organize summaries into fields such as goals, constraints, progress, decisions, and next steps. We call these fields \emph{slots}. Each slot describes one aspect of the task state and can be generated directly from the shared history. In our preliminary study, slots generated without using other slots' outputs still preserve the necessary information, showing the potential for independent slot generation. Furthermore, reversing the slot generation order has little effect on the retained information. These findings suggest that slots can serve as units for independent and parallel generation, with verification and repair performed within each slot.

We introduce \textbf{SpecCompact}, a framework for accelerating agent context compaction through slot-parallel semantic speculation. A fine-tuned small model drafts each slot from the shared history and the content already accepted in that slot. The target model checks whether each proposed line is supported by the history and consistent with the accepted content. It accepts the line or generates a local replacement. These draft--verify--repair loops run in parallel, and their outputs are combined into a structured summary. We also introduce system-level optimizations that reduce execution overhead and further accelerate compaction.

Our contributions are as follows:
\begin{itemize}
    \item We show that summary slots can retain relevant information without conditioning on other slots' outputs, supporting their use as units of parallel compaction.

    \item We introduce SpecCompact, which combines training, parallel drafting, and line-wise semantic verification with local repair. We further optimize its execution to reduce repeated context processing and model calls.

    \item We demonstrate $2.04$--$5.63\times$ compaction speedups over Hermes-style summarization, alongside downstream task evaluations and a native Hermes user study.
\end{itemize}

\section{Background and Motivation}
\subsection{Background}
Speculative decoding uses a small draft model to propose several tokens, which a larger target model verifies in a single forward pass~\citep{leviathan2023,chen2023specsampling}. The accepted prefix is kept, and the first rejected token is replaced before drafting resumes. This allows each target-model call to produce multiple tokens while remaining lossless relative to target-only decoding.

Semantic speculation relaxes the lossless requirement by evaluating draft content based on its meaning. Related methods such as SpecReasons~\citep{specreason} and DREAM-R~\citep{dreamr} apply this idea to reasoning steps for a lossy but much faster speculation inference. Like speculative decoding, these methods use a small model to draft a token sequence. Instead of verifying individual tokens, the target model evaluates the sequence as a complete reasoning step and outputs an accept-or-reject decision. In context compaction, summaries with different wording can preserve the same information needed for later actions. We therefore adopt semantic speculation for compaction, judging drafts by whether they preserve the information needed for later actions.

\subsection{Motivation}
Agent frameworks already use structured templates for context compaction. As shown in Figure~\ref{fig:slot-motivation}(a), OpenClaw and Hermes organize summaries into fields~\citep{openclaw,hermesagent}. In our context, we call these fields \emph{slots}. Each slot describes one aspect of the task state, such as the goal or current progress. Slots provide a natural unit for generation, but generating all slots as one sequence still requires later slots to wait for earlier ones. We therefore investigate whether this sequential dependency is necessary.

We compare monolithic generation with independent slot generation on 100 trajectories. Monolithic generation produces all six slots in a single output, while independent generation produces each slot from the same history without access to other slots' outputs. We use an LLM judge to evaluate how well each slot preserves relevant information from the full history, with a score from 0 to 5 obtained by summing five binary criteria. As shown in Figure~\ref{fig:slot-motivation}(b), independent generation achieves a mean score of 4.73, compared with 4.44 for monolithic generation. The fraction of slots scoring at least 4 is 93\% and 85\%, respectively. Reversing the slot order for monolithic generation or the invocation order for independent generation changes the mean score by at most 0.03. These results suggest that the information needed for a slot can often be recovered from the history without conditioning on other slots. This motivates generating slots in parallel from the shared history.

\begin{observationbox}{1}
Slots generated without other slots' outputs can still preserve relevant information. This supports generating slots in parallel from the shared history.
\end{observationbox}

\section{SpecCompact}\label{sec:method}
\begin{figure}[t]
    \centering
    \includegraphics[width=1\linewidth]{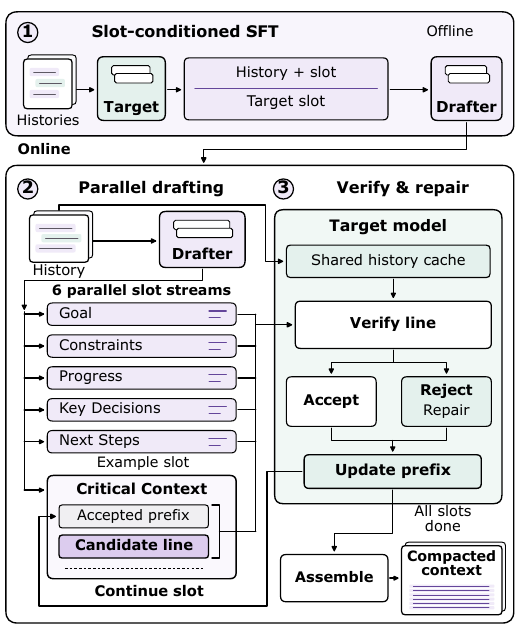}
    \caption{Overview of SpecCompact.}
    \label{fig:speccompact-system}
    \par\vspace{-10pt}
\end{figure}

Figure~\ref{fig:speccompact-system} provides an overview of SpecCompact. We first fine tune a small drafter on target-generated slot content. During compaction, the drafter generates slots in parallel from the shared history. The target model verifies each proposed line against the history and the accepted slot prefix, repairing rejected lines. Each slot continues from its updated prefix, and the slots are assembled into the compacted context once all are complete.

\subsection{Slot-Conditioned Drafting}
Based on our observation, we use slots as units of parallel generation. We follow OpenClaw's six-slot summary schema~\citep{openclaw}: Goal, Constraints and Preferences, Progress, Key Decisions, Next Steps, and Critical Context. Each slot has an instruction specifying its content and output format. Each request includes the shared conversation history and the full schema, but asks the drafter to generate only the requested slot. Figure~\ref{fig:drafter-prompt} provides the prompt template and slot contracts.

With the slot structure defined, we next fine-tune the drafter for slot-conditioned generation. We first use the target model with the generation prompt to produce supervised training data. We choose an independent small model as the drafter, using Qwen3-0.6B~\citep{yang2025qwen3technicalreport} in our experiments. Unlike vanilla speculative decoding, semantic speculation does not require the drafter and target to share a vocabulary. We align the drafter with the target's slot generation through supervised fine-tuning (SFT). For each training example, we minimize the cross-entropy loss:
\begin{equation}
    \mathcal{L}_{\mathrm{SFT}}(\theta)
    = -\sum_{j=1}^{L_s}
    \log q_{\theta}\!\left(y_{s,j} \mid H, c_s, y_{s,<j}\right),
    \label{eq:slot-sft}
\end{equation}
where $q_{\theta}$ is the drafter, $H$ is the history, $c_s$ is the slot instruction, and $y_s$ is the target-generated slot content represented as $L_s$ tokens under the drafter's tokenizer. Other objectives, such as KL divergence between aligned output distributions, could also be used. We use text-only supervision to avoid storing target probability distributions. We construct one training example for each history-slot pair, so a dataset of $T$ histories yields $6T$ SFT examples.

During compaction, we run a separate drafting stream for each slot. All streams use the same fine-tuned drafter and shared history, but maintain their own slot instructions and accepted prefixes. While SFT uses complete slot outputs, online drafting proceeds through line-level proposals. The drafter generates tokens autoregressively and submits each proposed line to the target model for verification.

For slot $s$, let $A_{s,t-1}$ denote the accepted prefix before step $t$, with $A_{s,0}$ initially empty. This prefix contains the lines already accepted or generated as replacements by the target model. Given the history $H$ and slot instruction $c_s$, the drafter $q_{\theta}$ proposes the next line:
\begin{equation}
    \tilde{\ell}_{s,t}
    \sim q_{\theta}\!\left(\cdot \mid H, c_s, A_{s,t-1}\right).
\end{equation}
The proposed line is added to the prefix only after verification. If accepted, it is appended unchanged. Otherwise, the target-generated replacement is appended instead. The updated prefix then conditions subsequent drafting, so the drafter continues from the content retained in the summary, including any target repairs. The history and slot instruction remain fixed throughout this process. Each slot depends only on its own prefix, allowing the six drafting streams to run in parallel.

\subsection{Semantic Verification}
As in DREAM-R~\citep{dreamr}, we use binary semantic verification. For each proposed line $\tilde{\ell}_{s,t}$, the target model checks whether its content is supported by the history $H$ and consistent with the accepted slot prefix $A_{s,t-1}$. The verification prompt includes the history, slot instruction, accepted prefix, and proposed line. It asks the target to return \texttt{positive} when both conditions hold, and \texttt{negative} otherwise. Figure~\ref{fig:target-verification-prompt} provides the verification prompt. We reduce the cost of repeatedly processing the history through shared-prefix caching, as discussed in the Section~\ref{sec:optimization}.

The target generates one judgment token, from which we obtain the top-$k$ token probabilities. Let $\hat{p}^{+}_{s,t}$ and $\hat{p}^{-}_{s,t}$ denote the probability estimates for \texttt{positive} and \texttt{negative}. We compute the normalized acceptance score as
\begin{equation}
    r_{s,t}
    = \frac{\hat{p}^{+}_{s,t}}
    {\hat{p}^{+}_{s,t}+\hat{p}^{-}_{s,t}},
    \label{eq:verification-score}
\end{equation}
and accept the line when $r_{s,t} \geq \eta$. Although we define only two labels, \texttt{positive} and \texttt{negative}, each may have multiple token forms in the target tokenizer. For example, \texttt{positive} may start with a capital letter or include a leading space. For each label, we use the highest probability among its returned token variants. If a label is missing from the returned top-$k$ tokens, we first compute the remaining probability mass as $1-\sum_{i=1}^{k}p_i$, where $p_i$ is the probability of the $i$-th returned token. We then compare this value with the smallest returned token probability and use the smaller value as the substitute probability. 

When a line is rejected, the target generates a replacement conditioned on the same history, slot instruction, and accepted prefix. The rejected line is excluded from this input. Given an accepted draft line or a valid target replacement, we update the slot prefix as
\begin{equation}
    A_{s,t} =
    \begin{cases}
        A_{s,t-1} \oplus \tilde{\ell}_{s,t}, & \text{if accepted}, \\
        A_{s,t-1} \oplus \ell^{\mathrm{T}}_{s,t}, & \text{if rejected},
    \end{cases}
    \label{eq:slot-prefix-update}
\end{equation}
where $\ell^{\mathrm{T}}_{s,t}$ is the target-generated replacement and $\oplus$ denotes concatenation. Drafting then continues from the updated prefix without regenerating previously accepted content.

\subsection{Efficient Slot-Parallel Execution}
\label{sec:optimization}

\begin{figure*}[t]
    \centering
    \includegraphics[width=1\linewidth]{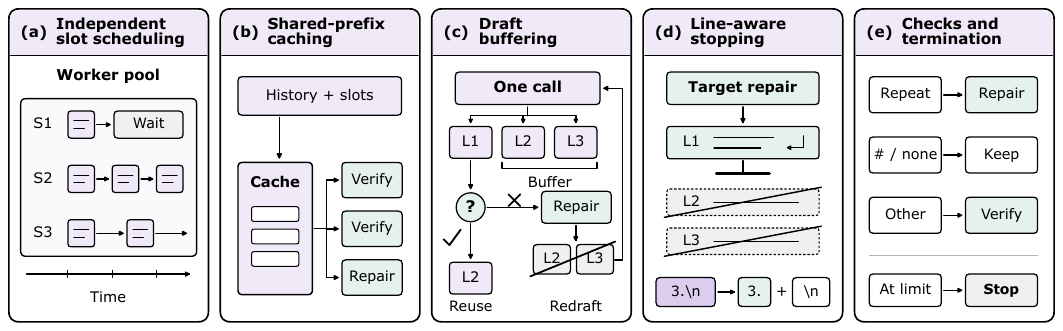}
    \caption{Execution optimizations in SpecCompact: independent slot scheduling, shared-prefix caching, draft buffering, line-aware stopping, and lightweight checks with termination rules.}
    \label{fig:optimization}
    \par\vspace{-10pt}
\end{figure*}

Generating slots in parallel does not by itself make line-wise speculation efficient. Each slot repeatedly invokes the drafter and target, and these calls include the same long history. Without reuse, even a one-token verification response requires another expensive prefill. Requesting each draft line separately also adds model-call overhead, while repair generation may decode additional lines that are never used. We address these costs through the following optimizations.

\paragraph{Independent slot scheduling.} We run the six slot streams concurrently, each with its own accepted prefix and sequential proposal, verification, and repair loop. While one slot waits for target verification or repair, other slots can continue drafting or submit their own verification requests. This allows work on the drafter and target to overlap and enables server-side batching across slots, reducing serial waiting. After all streams finish, we assemble their outputs in schema order.

\paragraph{Shared-prefix caching.} Each verification or repair call includes the full conversation history. Repeatedly prefilling this long input can be expensive even when the target generates only a short response. We then organize target requests with a common prefix containing the history and all six slot contracts. Slot instructions and the accepted prefix follow this shared portion, with the proposed line also included for verification. This keeps the leading tokens unchanged across slots and steps.

\paragraph{Draft buffering.} Calling the drafter separately for every line adds request overhead. We therefore let the drafter generate multiple lines in a single call. We send the first line to the target for verification and store the remaining complete lines in a buffer. If a line is accepted unchanged, the next buffered line can be checked without another draft call. Reuse requires the prefix used to generate that line to exactly match the current accepted prefix $A_{s,t-1}$. If a target repair changes the prefix, we discard the remaining buffered lines and generate a new continuation from the repaired prefix. Each buffered proposal undergoes the same line-wise checks before acceptance.

\paragraph{Line-aware stopping.} Each repair operation after rejection replaces one rejected line. However, the target may continue generating additional lines that are not used. We configure the target server to stop generation at the next line break, subject to the token limit, avoiding computation on these unused lines. To detect line breaks, we include all tokens ending in a newline in the stop set. Some of these tokens also contain text, so we retain that text when stopping. We preserve line and paragraph breaks when constructing the next input and keep any final content generated before end-of-sequence. The tokenizer and stop-token set are reused across calls to avoid repeated setup.

\paragraph{Checks and termination.}
In practice, the drafter sometimes produces repeated or abnormally long lines. We check whether a proposed line duplicates an accepted line or exceeds a configured character limit. Such lines go directly to target repair, avoiding an additional semantic verification call before replacement. We also cap the tokens generated per request and the lines and tokens retained per slot, preventing repetitive generation from continuing indefinitely. A slot stops when its budget is reached or a proposed addition would exceed it. Target replacements have the same checks and budget validation without further semantic verification. If a replacement is empty or fails these checks, we stop the slot and retain its accepted prefix. Structured summaries may produce standalone headings. After the length and duplicate checks, we directly accept lines beginning with \texttt{\#} and recognized empty-slot markers such as \texttt{(none)}, avoiding semantic verification calls for these standard output forms. Blank lines are retained as formatting. After assembly, we check the total summary token budget and raise an error if it is exceeded.

\begin{table*}[!t]
\centering
\small
\setlength{\tabcolsep}{1.3pt}
\renewcommand{\arraystretch}{1.12}
\caption{Agent task performance (\%). Full History is a reference without compaction. Bold marks the best compaction result. Gray subscripts show 95\% confidence intervals.}
\label{tab:eval:agent}
\resizebox{\textwidth}{!}{%
\begin{tabular}{ll|cc|cc|c}
\toprule
\multirow{2}{*}{\textbf{Models}} & \multirow{2}{*}{\textbf{Methods}} & \multicolumn{2}{c|}{\textbf{AppWorld Normal}} & \multicolumn{2}{c|}{\textbf{AppWorld Challenge}} & \textbf{$\tau^3$} \\
\cmidrule(lr){3-4}\cmidrule(lr){5-6}\cmidrule(lr){7-7}
 &  & \textbf{TGC} $\uparrow$ & \textbf{SGC} $\uparrow$ & \textbf{TGC} $\uparrow$ & \textbf{SGC} $\uparrow$ & \textbf{Success} $\uparrow$ \\
\midrule
\multirow{4}{*}{\makecell{Gemma4-\\31B-IT}} & Full History & 62.50\CI{51.79}{72.62} & 46.43\CI{33.93}{58.93} & 53.24\CI{45.80}{60.43} & 41.73\CI{33.81}{49.64} & 52.08\CI{45.00}{58.75} \\
 & OpenClaw & \textbf{58.93}\CI{47.62}{69.64} & \textbf{44.64}\CI{32.14}{57.14} & 46.52\CI{39.09}{53.72} & \textbf{33.09}\CI{25.18}{41.01} & 47.50\CI{39.17}{55.42} \\
 & Hermes & 57.74\CI{47.02}{67.86} & 39.29\CI{26.79}{51.79} & \textbf{47.24}\CI{40.05}{54.44} & 32.37\CI{25.18}{40.29} & \textbf{50.00}\CI{42.08}{57.92} \\
 & \cellcolor{SpecCompactFill}\textbf{SpecCompact} & \cellcolor{SpecCompactFill}54.76\CI{44.05}{65.48} & \cellcolor{SpecCompactFill}37.50\CI{25.00}{50.00} & \cellcolor{SpecCompactFill}43.17\CI{35.97}{50.12} & \cellcolor{SpecCompactFill}28.06\CI{20.86}{35.25} & \cellcolor{SpecCompactFill}47.50\CI{40.00}{55.00} \\
\midrule
\multirow{4}{*}{\makecell{Qwen3.5-\\122B-A10B}} & Full History & 40.48\CI{29.76}{51.19} & 26.79\CI{16.07}{37.50} & 20.62\CI{15.35}{26.14} & 8.63\CI{4.32}{13.67} & 67.92\CI{63.33}{72.08} \\
 & OpenClaw & 28.57\CI{19.05}{38.69} & 14.29\CI{5.36}{23.21} & \textbf{17.75}\CI{12.71}{23.26} & \textbf{8.63}\CI{4.32}{13.67} & \textbf{65.83}\CI{60.83}{70.83} \\
 & Hermes & 28.57\CI{19.05}{38.69} & 14.29\CI{5.36}{23.21} & 14.39\CI{9.59}{19.42} & 7.19\CI{2.88}{11.51} & 63.75\CI{58.33}{68.75} \\
 & \cellcolor{SpecCompactFill}\textbf{SpecCompact} & \cellcolor{SpecCompactFill}\textbf{33.33}\CI{23.21}{44.05} & \cellcolor{SpecCompactFill}\textbf{21.43}\CI{10.71}{32.14} & \cellcolor{SpecCompactFill}16.07\CI{11.27}{21.10} & \cellcolor{SpecCompactFill}6.47\CI{2.88}{10.79} & \cellcolor{SpecCompactFill}64.58\CI{59.17}{69.58} \\
\bottomrule
\end{tabular}%
}
\par\vspace{-10pt}
\end{table*}

\section{Experiments}
We evaluate SpecCompact with Gemma4-31B-IT~\citep{gemmateam2026gemma4technicalreport} and Qwen3.5-122B-A10B~\citep{qwen3.5} as target models, each paired with a separately fine-tuned Qwen3-0.6B drafter~\citep{yang2025qwen3technicalreport}. We use AppWorld~\citep{trivedi2024appworldcontrollableworldapps} and $\tau^3$~\citep{shi2026tauknowledgeevaluatingconversationalagents} to evaluate whether compacted histories preserve the information needed for subsequent agent actions. AppWorld tests task completion through code and application APIs, while $\tau^3$ tests task completion through user interaction and tool use. We evaluate on the full AppWorld Test-Normal and Test-Challenge sets, and run three trials per task on an 80-task random selected subset of $\tau^3$ for each target model. We report task goal completion (TGC) and scenario goal completion (SGC) on AppWorld, and task success rate on $\tau^3$. Beyond agent tasks, we also evaluate long-context compaction on LongBench~\citep{bai2023longbench} and RULER~\citep{hsieh2024rulerwhatsrealcontext}. For LongBench, we use the GovReport and QMSum subsets and report ROUGE-L. For RULER, we evaluate full set of five task types at context lengths of 8k, and 16k using the official task-specific matching metrics. To measure compaction latency and throughput, we construct smaller subsets from AppWorld, $\tau^3$, and LongBench instead of using the complete sets.

We construct about 50K training data from AppWorld, $\tau^3$, and LiveClawBench~\citep{long2026liveclawbenchbenchmarkingllmagents}, using each target model to generate slot-level supervision. Training and validation examples are separated by trajectory ID. We fine-tune each drafter for one epoch with LoRA~\citep{hu2021loralowrankadaptationlarge} applied to all linear layers, using rank 16, scaling factor 32, and zero dropout. Training uses AdamW~\citep{loshchilov2019decoupledweightdecayregularization}, with a learning rate of $10^{-4}$, cosine decay, a warmup ratio of 0.05, zero weight decay, and gradient clipping at 1.0. We use a per-device batch size of 2 with 16 gradient accumulation steps, BF16 precision, a maximum sequence length of 16,384 tokens, and random seed 42. The cross-entropy loss is computed only over the target slot content.

During agent evaluation, we compact the interaction history every eight agent model calls, retaining system and setup messages and the current input. New messages accumulate until the next compaction. For latency experiments, we serve the models with vLLM~\citep{kwon2023efficientmemorymanagementlarge} on NVIDIA RTX PRO 6000 Blackwell GPUs. Gemma4 runs in BF16 on one GPU, while Qwen3.5 runs in FP8 on two GPUs. SpecCompact uses one additional GPU for the drafter and runs six slots concurrently with an acceptance threshold of $\eta=0.7$. Drafting and repair allow up to 64 new tokens per request, while verification uses one token. Each slot is limited to 12 lines and 320 tokens. The summary budget is 2,048 tokens for SpecCompact and the baselines.

\subsection{Agent Task Performance}
Table~\ref{tab:eval:agent} reports downstream task performance on AppWorld and $\tau^3$. The effect of SpecCompact varies across target models. Compared with Hermes, SpecCompact reduces Gemma4's TGC by 2.98 and 4.07 percentage points on AppWorld Normal and Challenge, respectively, and reduces $\tau^3$ success by 2.50 points. For Qwen3.5, SpecCompact increases TGC by 4.76 and 1.68 points on the two AppWorld splits and increases $\tau^3$ success by 0.83 points, although Challenge SGC decreases by 0.72 points. SpecCompact therefore offers faster compaction with a model-dependent effect on downstream performance.

\begin{table*}[!t]
\centering
\small
\setlength{\tabcolsep}{4.5pt}
\renewcommand{\arraystretch}{1.12}
\caption{Long-context performance on LongBench (ROUGE-L) and RULER at 8k and 16k (official matching scores), scaled to 100. Execution failures are scored as zero. Bold marks the best compaction result.}
\label{tab:eval:long_context}
\resizebox{\textwidth}{!}{%
\begin{tabular}{ll|cc|cc}
\toprule
\multirow{2}{*}{\textbf{Models}}
& \multirow{2}{*}{\textbf{Methods}}
& \multicolumn{2}{c|}{\textbf{LongBench}}
& \multicolumn{2}{c}{\textbf{RULER}} \\
\cmidrule(lr){3-4}\cmidrule(lr){5-6}
& & \textbf{GovReport} $\uparrow$
& \textbf{QMSum} $\uparrow$
& \textbf{8k} $\uparrow$
& \textbf{16k} $\uparrow$ \\
\midrule
\multirow{3}{*}{\makecell{Gemma4-\\31B-IT}}
& Full History
& 28.36\CI{27.73}{29.02}
& 24.34\CI{23.28}{25.41}
& 91.72\CI{88.77}{94.62}
& 92.12\CI{89.22}{95.02} \\
& Target-self
& \textbf{25.05}\CI{24.51}{25.59}
& \textbf{18.35}\CI{17.61}{19.10}
& \textbf{71.00}\CI{67.68}{74.67}
& \textbf{71.02}\CI{67.72}{74.70} \\
& \cellcolor{SpecCompactFill}\textbf{SpecCompact}
& \cellcolor{SpecCompactFill}22.28\CI{21.10}{23.37}
& \cellcolor{SpecCompactFill}17.52\CI{16.51}{18.52}
& \cellcolor{SpecCompactFill}66.50\CI{63.67}{69.70}
& \cellcolor{SpecCompactFill}68.55\CI{65.32}{71.87} \\
\midrule
\multirow{3}{*}{\makecell{Qwen3.5-\\122B-A10B}}
& Full History
& 31.16\CI{30.53}{31.82}
& 22.98\CI{22.03}{23.96}
& 90.92\CI{87.63}{93.82}
& 90.92\CI{87.63}{94.17} \\
& Target-self
& \textbf{28.25}\CI{27.71}{28.79}
& \textbf{18.54}\CI{17.86}{19.25}
& 72.98\CI{69.33}{76.67}
& 72.15\CI{68.87}{75.80} \\
& \cellcolor{SpecCompactFill}\textbf{SpecCompact}
& \cellcolor{SpecCompactFill}27.66\CI{27.05}{28.26}
& \cellcolor{SpecCompactFill}18.05\CI{17.24}{18.87}
& \cellcolor{SpecCompactFill}\textbf{74.23}\CI{70.55}{77.93}
& \cellcolor{SpecCompactFill}\textbf{72.60}\CI{68.97}{76.27} \\
\bottomrule
\end{tabular}%
}
\par\vspace{-10pt}
\end{table*}
\subsection{Long-Context Performance}

Table~\ref{tab:eval:long_context} evaluates compaction on LongBench and RULER at 8k and 16k. Target-self uses the target model to generate each slot directly, without a drafter or semantic verification. On Qwen3.5, SpecCompact scores within 0.6 points of Target-self on both LongBench datasets and achieves higher scores at both RULER lengths. On Gemma4, SpecCompact scores 0.83--4.50 points below Target-self across the four settings.

\begin{table*}[!t]
\centering
\small
\setlength{\tabcolsep}{3pt}
\renewcommand{\arraystretch}{1.12}
\caption{Compaction latency and speedup on 20 matched inputs per workload. Speedup is relative to Hermes. Bold marks the best result within each target. SpecCompact uses an additional drafter GPU.}
\label{tab:eval:system}
\resizebox{\textwidth}{!}{%
\begin{tabular}{ll|cc|cc|cc|cc}
\toprule
\multirow{3}{*}{\textbf{Models}}
& \multirow{3}{*}{\textbf{Methods}}
& \multicolumn{4}{c|}{\textbf{AppWorld}}
& \multicolumn{2}{c|}{\multirow{2}{*}{\textbf{$\tau^3$}}}
& \multicolumn{2}{c}{\multirow{2}{*}{\textbf{LongBench 16k}}} \\
\cmidrule(lr){3-6}
& & \multicolumn{2}{c|}{\textbf{Normal}}
& \multicolumn{2}{c|}{\textbf{Challenge}}
& \multicolumn{2}{c|}{}
& \multicolumn{2}{c}{} \\
\cmidrule(lr){3-4} \cmidrule(lr){5-6}
\cmidrule(lr){7-8} \cmidrule(lr){9-10}
& & Time (s) $\downarrow$ & Speedup $\uparrow$
& Time (s) $\downarrow$ & Speedup $\uparrow$
& Time (s) $\downarrow$ & Speedup $\uparrow$
& Time (s) $\downarrow$ & Speedup $\uparrow$ \\
\midrule
\multirow{3}{*}{\makecell{Gemma4-\\31B-IT}}
& Hermes
& 38.89 & 1.00$\times$
& 39.12 & 1.00$\times$
& 36.21 & 1.00$\times$
& 74.13 & 1.00$\times$ \\
& OpenClaw
& 16.44 & 2.37$\times$
& 18.70 & 2.09$\times$
& 13.19 & 2.75$\times$
& 63.55 & 1.17$\times$ \\
& \cellcolor{SpecCompactFill}\textbf{SpecCompact}
& \cellcolor{SpecCompactFill}\textbf{8.26}
& \cellcolor{SpecCompactFill}\textbf{4.71}$\times$
& \cellcolor{SpecCompactFill}\textbf{9.14}
& \cellcolor{SpecCompactFill}\textbf{4.28}$\times$
& \cellcolor{SpecCompactFill}\textbf{6.43}
& \cellcolor{SpecCompactFill}\textbf{5.63}$\times$
& \cellcolor{SpecCompactFill}\textbf{29.72}
& \cellcolor{SpecCompactFill}\textbf{2.49}$\times$ \\
\midrule
\multirow{3}{*}{\makecell{Qwen3.5-\\122B-A10B}}
& Hermes
& 18.03 & 1.00$\times$
& 16.98 & 1.00$\times$
& 17.81 & 1.00$\times$
& 29.43 & 1.00$\times$ \\
& OpenClaw
& 7.87 & 2.29$\times$
& 8.69 & 1.95$\times$
& \textbf{7.68} & \textbf{2.32}$\times$
& 24.62 & 1.20$\times$ \\
& \cellcolor{SpecCompactFill}\textbf{SpecCompact}
& \cellcolor{SpecCompactFill}\textbf{7.49}
& \cellcolor{SpecCompactFill}\textbf{2.41}$\times$
& \cellcolor{SpecCompactFill}\textbf{7.55}
& \cellcolor{SpecCompactFill}\textbf{2.25}$\times$
& \cellcolor{SpecCompactFill}8.71
& \cellcolor{SpecCompactFill}2.04$\times$
& \cellcolor{SpecCompactFill}\textbf{21.67}
& \cellcolor{SpecCompactFill}\textbf{1.36}$\times$ \\
\bottomrule
\end{tabular}%
}
\par\vspace{-10pt}
\end{table*}

\subsection{Compaction Efficiency}

Table~\ref{tab:eval:system} reports compaction latency on matched inputs. Relative to Hermes, SpecCompact achieves $4.28$--$5.63\times$ speedups on the agent workloads with Gemma4 and $2.04$--$2.41\times$ with Qwen3.5. On LongBench inputs of approximately 16k tokens, the speedups are $2.49\times$ and $1.36\times$, respectively. Three factors help explain the smaller speedups with Qwen3.5. First, Qwen3.5's baseline summarization is already faster. Second, the larger model-size gap makes it harder for the 0.6B drafter to match Qwen3.5's behavior, with a draft acceptance rate 19.6 percentage points lower than with Gemma4. Third, our vLLM deployment provides limited prefix-cache reuse for Qwen3.5's hybrid attention. Standard attention can reuse cached prefix KV states, whereas Qwen3.5's linear-attention layers also need to restore recurrent states from checkpoints saved at suitable positions. Matching text prefixes alone therefore do not guarantee cache reuse. SpecCompact also outperforms OpenClaw in seven of the eight settings. The exception is Qwen3.5 on $\tau^3$, where SpecCompact takes 8.71 seconds compared with OpenClaw's 7.68 seconds.

\subsection{Effect of System Optimizations}

\begin{table}[h]
\centering
\small
\setlength{\tabcolsep}{8.3pt}
\renewcommand{\arraystretch}{1.12}
\caption{Combined effect of system optimizations on 20 matched inputs per workload. Both versions use slot-parallel execution. Times are in seconds.}
\label{tab:eval:optimizations}
\resizebox{\columnwidth}{!}{%
\begin{tabular}{l|ccc}
\toprule
\textbf{Benchmark}
& \textbf{Before} $\downarrow$
& \textbf{After} $\downarrow$
& \textbf{Speedup} $\uparrow$ \\
\midrule
\multicolumn{4}{c}{\textbf{Gemma4-31B-IT}} \\
\midrule
AppWorld Normal & 20.17
& \cellcolor{SpecCompactFill}\textbf{8.26}
& \cellcolor{SpecCompactFill}\textbf{2.44}$\times$ \\
AppWorld Challenge & 21.56
& \cellcolor{SpecCompactFill}\textbf{9.14}
& \cellcolor{SpecCompactFill}\textbf{2.36}$\times$ \\
$\tau^3$ & 12.07
& \cellcolor{SpecCompactFill}\textbf{6.43}
& \cellcolor{SpecCompactFill}\textbf{1.88}$\times$ \\
\midrule
\multicolumn{4}{c}{\textbf{Qwen3.5-122B-A10B}} \\
\midrule
AppWorld Normal & 9.55
& \cellcolor{SpecCompactFill}\textbf{7.49}
& \cellcolor{SpecCompactFill}\textbf{1.28}$\times$ \\
AppWorld Challenge & 11.43
& \cellcolor{SpecCompactFill}\textbf{7.55}
& \cellcolor{SpecCompactFill}\textbf{1.51}$\times$ \\
$\tau^3$ & 10.38
& \cellcolor{SpecCompactFill}\textbf{8.71}
& \cellcolor{SpecCompactFill}\textbf{1.19}$\times$ \\
\bottomrule
\end{tabular}%
}
\par\vspace{-10pt}
\end{table}

Table~\ref{tab:eval:optimizations} compares SpecCompact before and after the system optimizations. Both implementations already generate slots concurrently, allowing us to examine the additional gains from optimizing their execution. The optimizations reduce compaction latency by $1.88$--$2.44\times$ for Gemma4 and $1.19$--$1.51\times$ for Qwen3.5. For example, Gemma4's latency on AppWorld Normal falls from 20.17 to 8.26 seconds. These gains highlight the value of reducing repeated context processing, model calls, and unnecessary decoding beyond slot parallelism alone. The table measures the combined effect of the optimizations rather than the contribution of each component. The full task evaluation uses the original implementation and has not been repeated with all system optimizations.

\subsection{User Study}

To evaluate user experience in a real agent setting, we integrate SpecCompact into the Hermes codebase and deploy the system with Gemma4 on four NVIDIA RTX PRO 6000 GPUs. We construct three scenarios from our own workflows, covering coding, math, and multitask question answering. For each scenario, we run Hermes in advance and pause at a fixed checkpoint to save a context snapshot. Users trigger compaction from the same snapshot with each method and select which feels faster without knowing the method identities. We also record the actual compaction time in the background. After compaction, we ask the agent to produce a summary and collect a separate blind preference between the two outputs.

According to the collected compaction data, SpecCompact reduces median compaction latency from 63.74 to 7.28 seconds, with a median paired speedup of $8.68\times$. These measurements show that the compaction speedups also carry over to the native agent deployment. Among 28 summary-preference responses, 11 (39.3\%) favor ours, 8 (28.6\%) favor Hermes, 8 (28.6\%) rate them similarly, and 1 (3.6\%) cannot judge.

\section{Related Work}
\subsection{Agent Context Management}
LLM agents manage growing histories through external memory and context compression. Memory systems store and retrieve past information~\citep{memgpt,amem}, while prompt and KV-cache compression reduce prompt length or cache size~\citep{llmlingua,recomp,snapkv}. For long-running agents, context compaction summarizes interaction histories into a shorter working context. Existing methods improve compression guidelines~\citep{acon}, periodically summarize trajectories~\citep{resum}, or train agents to manage their context~\citep{mem1,supo,contextfolding,agentfold}. Other work improves how compaction is executed. Slipstream runs compaction alongside continued agent execution and uses subsequent steps to validate summaries~\citep{slipstream}. Parallel compaction summarizes input blocks concurrently~\citep{parallelcompaction}. SpecCompact accelerates the generation and verification of structured summaries through slot-parallel semantic speculation.

\subsection{Speculative Decoding and Reasoning}
Speculative decoding uses a small model to draft tokens and a target model to verify them in parallel, preserving the target model's output distribution~\citep{leviathan2023,chen2023specsampling}. Semantic speculation relaxes this requirement. SpecReason uses target-model judgments to verify complete reasoning steps~\citep{specreason}. DREAM-R further aligns the drafter through reinforcement learning and parallelizes drafting and verification~\citep{dreamr}. This approach fits context compaction, where different summaries can preserve the information needed for later actions. SpecCompact adapts semantic speculation to compaction through parallel slot generation and line-wise verification.

\section{Discussion and Limitations}
\paragraph{Limitations.} SpecCompact focuses on the trade-off between accuracy and compaction speed, with an emphasis on the framework and its system implementation. It does not offer a consistent accuracy advantage over target-model summarization. We believe better drafter--target alignment, such as the reinforcement learning approach used in DREAM-R~\citep{dreamr}, could improve accuracy and provide a better trade-off.

\paragraph{Discussion.} We also find that slots differ in where they obtain information from the history. Progress and Decisions use evidence spread throughout the context, while Goal, Constraints, and Next Steps mainly rely on beginning or end. This suggests slot-conditioned sliding-window attention as a way to further accelerate drafting. We leave this direction to future work due to limited time and resources.

\section{Conclusion}
We presented SpecCompact for accelerating context compaction in LLM agents. SpecCompact combines slot-parallel drafting, line-wise semantic verification, and system optimizations to reduce compaction latency. Experiments across agent and long-context benchmarks demonstrate substantial speedups with accuracy trade-offs, while a native Hermes user study confirms the latency benefits.

\bibliography{custom}

\appendix
\section{Prompt Template}
\label{app:prompt}
\subsection{Drafter Prompt}
Figure~\ref{fig:drafter-prompt} presents the drafter prompt and the six slot contracts. Each request includes the shared conversation history and the complete summary schema, together with the requested slot's title, contract, and output format. The drafter generates only the requested slot's content.
\begin{figure*}[t]
\begin{promptbox}{Drafter Prompt}
You are a context summarization assistant. Your task is to read a conversation between a user and an AI coding assistant, then produce structured summary content following the requested format.

Do NOT continue the conversation. Do NOT respond to any questions in the conversation. ONLY output the requested summary content.

The messages below are a conversation to summarize. The next request will ask for one section of a structured context checkpoint summary that another LLM will use to continue the work.

The complete OpenClaw-style summary schema contains these sections:\\
\promptvar{slot_lines}

Conversation:\\
\texttt{<conversation>}\\
\promptvar{trajectory}\\
\texttt{</conversation>}

Keep the requested section concise and consistent with how it would appear inside the complete OpenClaw-style summary. Preserve exact file paths, function names, and error messages.

Slot request:\\
Generate only the \texttt{\char0}\promptvar{slot}\texttt{\char0} slot as one section of the complete OpenClaw-style summary.\\
Slot title: \promptvar{slot_title}\\
Slot contract: \promptvar{slot_contract}

Use this exact slot-body format:\\
\promptvar{slot_format}

Return only the slot content, not JSON.\\
Do not include the top-level slot heading.\\
Nested headings required by the slot contract are allowed.
\end{promptbox}

\vspace{8pt}

\begin{promptbox}{Slot Contracts}
\slotname{goal}\\
What is the user trying to accomplish? Can be multiple items if the session covers different tasks.

\slotname{constraints_preferences}\\
Any constraints, preferences, or requirements mentioned by the user. Return (none) if none were mentioned.

\slotname{progress}\\
Use the OpenClaw Progress substructure: Done contains completed tasks/changes; In Progress contains current work; Blocked contains issues preventing progress, if any.

\slotname{key_decisions}\\
Decisions with brief rationale, using the pattern \texttt{**Decision**}: rationale.

\slotname{next_steps}\\
Ordered list of what should happen next.

\slotname{critical_context}\\
Any data, examples, or references needed to continue. Return (none) if not applicable. Preserve exact file paths, function names, and error messages.
\end{promptbox}

\caption{The drafter prompt template and the contract for each slot.}
\label{fig:drafter-prompt}
\end{figure*}

\subsection{Target Prompt}
Figure~\ref{fig:target-verification-prompt} shows the target verification prompt and its message structure. The full conversation history and all six slot contracts form a shared prefix, followed by the verification request for a proposed line. The target returns \texttt{positive} only if the line is supported by the history and consistent with the accepted summary lines.

\begin{figure*}[t]

\begin{promptbox}{Target Verification Prompt}
You are verifying one line of a context-compaction summary.\\
Reply 'positive' ONLY if the proposed line is factually supported by the trajectory AND coherent with the accepted summary lines so far.\\
Reply 'negative' if it contains any factual error, unsupported claim, contradiction, or content promoted beyond what the trajectory confirms.\\
Reply with exactly one word: positive or negative.

Summary slot: \promptvar{slot} -{}- \promptvar{slot_contract}

Accepted summary lines so far:\\
\promptvar{accepted_lines}

Proposed next line:\\
\promptvar{proposed_line}
\end{promptbox}

\vspace{10pt}

\begin{lstlisting}[style=messages]
[
    {
        "role": "user",
        "content": shared_prefix_with_full_history_and_all_six_slot_contracts,
    },
    {
        "role": "user",
        "content": verification_prompt_above,
    },
]
\end{lstlisting}

\caption{Target verification prompt (top) and the message structure with a shared prefix (bottom).}
\label{fig:target-verification-prompt}
\end{figure*}


\end{document}